\documentclass[11pt]{article} 

\usepackage[utf8]{inputenc} 

\usepackage{geometry} 
\usepackage{graphicx}
\usepackage{subfigure}
\usepackage{amssymb}
\usepackage{multirow}
\usepackage{hyperref}
\usepackage{amsmath}

 \usepackage{float}

\title{Evolutionary Multimodal Optimization: A Short Survey}
\author{Ka-Chun Wong (Department of Computer Science, University of Toronto)}

\begin{document}
\maketitle

Real world problems always have different multiple solutions. For
instance, optical engineers need to tune the recording parameters to get
as many optimal solutions as possible for multiple trials in the
varied-line-spacing holographic grating design problem. Unfortunately,
most traditional optimization techniques focus on solving for a single
optimal solution. They need to be applied several times; yet all
solutions are not guaranteed to be found. Thus the multimodal
optimization problem was proposed. In that problem, we are interested in
not only a single optimal point, but also the others. With strong
parallel search capability, evolutionary algorithms are shown to be
particularly effective in solving this type of problem. In particular,
the evolutionary algorithms for multimodal optimization usually not only
locate multiple optima in a single run, but also preserve their
population diversity throughout a run, resulting in their global
optimization ability on multimodal functions. In addition, the
techniques for multimodal optimization are borrowed as diversity
maintenance techniques to other problems. In this chapter, we describe and review the
state-of-the-arts evolutionary algorithms for multimodal optimization in terms of methodology, benchmarking, and application.

\section{Introduction and Background}

Since genetic algorithm was proposed by John H. Holland \cite{holland:GA} in the early 1970s, 
researchers have been 
exploring the power of evolutionary algorithms \cite{wong2014herd}. For instance, 
biological pattern discovery \cite{springerlink:10.1007/s00500-011-0692-5} and computer vision \cite{wu2010}. 
In particular, its function optimization capability was highlighted \cite{goldberg:book} because of its high adaptability to different non-convex function landscapes, to which we cannot apply traditional optimization techniques. 

Real world problems always have different multiple solutions \cite{gecco2010,pmid25192742}. For instance, optical engineers need to tune the recording parameters to get as many optimal solutions as possible for multiple trials in the varied-line-spacing holographic grating design problem because the design constraints are too difficult to be expressed and solved in mathematical forms \cite{opticalApp}. Unfortunately, most traditional optimization techniques focus on solving for a single optimal solution. They need to be applied several times; yet all solutions are not guaranteed to be found. Thus the multimodal optimization problem was proposed. In that problem, we are interested in not only a single optimal point, but also the others. Given an objective function, an algorithm is expected to find all optimal points in a single run. With strong parallel search capability, evolutionary algorithms are shown to be particularly effective in solving this type of problem \cite{goldberg:book}: Given function $f:\mathbb{X} \rightarrow \mathbb{R}$, we would like to find all global and local maxima (or minima) of $f$ in a single run. Although the objective is clear, it is not easy to be satisfied in practice because some problems may have too many optima to be located. Nonetheless, it is still of great interest to researchers how these problems are going to be solved because the algorithms for multimodal optimization usually not only locate multiple optima in a single run, but also preserve their population diversity throughout a run, resulting in their global optimization ability on multimodal functions. Moreover, the techniques for multimodal optimization are usually borrowed as diversity maintenance techniques to other problems \cite{pmid23814189,pmid24389653}.

\section{Problem Definition}
The multimodal optimization problem definition depends on the type of optimization (minimization or maximization). They are similar in principle and defined as follows:

\subsection{Minimization}
In this problem, given $f$:$\mathbb{X} \rightarrow \mathbb{R}$, we would like to find all global and local minimums of $f$ in a single run.
\\
\\
\textbf{Definition 1} Local Minimum \cite{W2007GOEB}: A (local) minimum $\hat{x_l} \in \mathbb{X}$ of one (objective) function $f$:$\mathbb{X}\rightarrow\mathbb{R}$ is an input element with $f(\hat{x_l}) \leq f(x)$ for all $x$ neighboring $\hat{x_l}$. If $\mathbb{X} \in \mathbb{R}^N$, we can write: $\forall \hat{x_l}$ $\exists \epsilon > 0$ : $f(\hat{x_l}) \leq f(x)$ $\forall x \in \mathbb{X}, |x-\hat{x_l}|<\epsilon$.
\\
\\
\textbf{Definition 2} Global Minimum \cite{W2007GOEB}: A global minimum $\hat{x_g} \in \mathbb{X}$ of one (objective) function $f$:$\mathbb{X}\rightarrow\mathbb{R}$ is an input element with $f(\hat{x_g}) \leq f(x)$ $\forall x \in \mathbb{X}$.

\subsection{Maximization}
In this problem, given $f$:$\mathbb{X} \rightarrow \mathbb{R}$, we would like to find all global and local maximums of $f$ in a single run.
\\
\\
\textbf{Definition 3} Local Maximum \cite{W2007GOEB}: A (local) maximum $\hat{x_l} \in \mathbb{X}$ of one (objective) function $f$:$\mathbb{X}\rightarrow\mathbb{R}$ is an input element with $f(\hat{x_l}) \geq f(x)$ for all $x$ neighboring $\hat{x_l}$. If $\mathbb{X} \in \mathbb{R}^N$, we can write: $\forall \hat{x_l}$ $\exists \epsilon > 0$ : $f(\hat{x_l}) \geq f(x)$ $\forall x \in \mathbb{X}, |x-\hat{x_l}|<\epsilon$.
\\
\\
\textbf{Definition 4} Global Maximum \cite{W2007GOEB}: A global maximum $\hat{x_g} \in \mathbb{X}$ of one (objective) function $f$:$\mathbb{X}\rightarrow\mathbb{R}$ is an input element with $f(\hat{x_g}) \geq f(x)$ $\forall x \in \mathbb{X}$.

\section{Methodology}
In the past literature, there are different evolutionary methods proposed for multimodal optimization. In this section, we discuss and categorize them into different methodologies.

\subsection{Preselection}
In 1970, the doctorial thesis by Cavicchio introduced different methods for genetic algorithms \cite{cavicchio1970adaptive}. In particular, the preselection scheme was proposed to maintain the population diversity. In this scheme, the children compete with their parents for survival. If a child has a fitness (measured by an objective function) higher than its parent, the parent is replaced by the child in the next generation.

\subsection{Crowding}
In 1975, the work by De Jong \cite{jong:crowding} introduced the crowding technique to increase the chance of locating multiple optima. In the crowding technique, each child is compared to a random sub-population of $cf$ members in the existing parent population ($cf$ means crowding factor). The parent member which is most similar to the child itself is selected (measured by a distance metric). If the child has a higher fitness than the parent member selected, then it replaces the parent member in the population. Besides genetic algorithm, Thomsen has also incorporated crowding techniques \cite{jong:crowding} into differential evolution (CrowdingDE) for multimodal optimization \cite{thomsen:crowdingde}. In his study, the crowding factor is set to the population size and Euclidean distance is adopted as the dissimilarity metric. The smaller the distance, the more similar they are and vice versa. Although an intensive computation is accompanied, it can effectively transform differential evolution into an algorithm specialized for multimodal optimization. In 2012, CrowdingDE has been investigated and extended by Wong et al , demonstrating competitive performance even when it is compared to the other state-of-the-arts methods \cite{wong2012evolutionary}. 

\subsection{Fitness Sharing}
In 1989, Goldberg and Richardson proposed a fitness-sharing niching technique as a diversity preserving strategy to solve the multimodal optimization problem \cite{goldberg:sharing}. They proposed a shared fitness function, instead of an absolute fitness function, to evaluate the fitness of a individual in order to favor the growth of the individuals which are distinct from the others. The shared fitness function is defined as follows:
\[
f'(x_i) = Shared\ Fitness = \frac{Actual\ Fitness}{Degree\ of\ Sharing} = \frac{f(x_i)}{\sum_{j=1}^N sh(d(x_i,x_j))}
\]
where $f'(x_i)$ is the shared fitness of the $i$th individual $x_i$; $f(x_i)$ is the actual fitness of the $i$th individual $x_i$; $d(x_i,x_j)$ is the distance function between the two individuals $x_i$ and $x_j$; $sh(d)$ is the sharing function. With this technique, a population can be prevented from the domination of a particular type of individuals. Nonetheless, a careful adjustment to the sharing function $sh(d)$ definition is needed because it relates the fitness domain $f(x_i)$ to the distance domain $d(x_i,x_j)$ which are supposed to be independent of each other.

\subsection{Species Conserving}
Species conserving genetic algorithm (SCGA) \cite{li:SCGA} is a technique for evolving parallel subpopulations for multimodal optimization. Before each generation starts, the algorithm selects a set of species seeds which can bypass the subsequent procedures and be saved into the next generation. The algorithm then divides a population into several species based on a dissimilarity measure. The fittest individual is selected as the species seed for each species. After the identification of species seeds, the population undergoes the usual genetic algorithm operations: selection, crossover, and mutation. As the operations may remove the survival of less fit species, the saved species seeds are copied back to the population at the end of each generation.

To determine the species seeds in a population, the algorithm first sorts the population in a decreasing fitness order. Once sorted, it picks up the fittest individual as the first species seed and forms a species region around it. The next fittest individual is tested whether it is located in a species region. If not, it is selected as a species seed and another species region is created around it. Otherwise, it is not selected. Similar operations are applied to the remaining individuals, which are subsequently checked against all existing species seeds.

To copy the species seeds back to the population after the genetic operations have been executed, the algorithms need to scan all the individuals in the current population and identify to which species they belong. Once it is identified, the algorithm replaces the worst individual (lowest fitness) with the species seed in a species. If no individuals can be found in a species for replacement, the algorithm replaces the worst and un-replaced individual in the whole population. In short, the main idea is to preserve the population diversity by preserving the fittest individual for each species.

\subsection{Covariance Matrix Adaptation}
Evolution strategy is an effective method for numerical optimization. In recent years, its variant CMA-ES (Covariance Matrix Adaptation Evolution Strategy) showed a remarkable success \cite{Hansen:2001:CDS:1108839.1108843}. To extend its capability, niching techniques have been introduced to cope with multimodal functions \cite{Shir:2010:ANR:1739146.1739150}. For instance, a concept called \textit{adaptive individual niche radius} has been proposed to solve the \textit{niche radius problem} commonly found in speciation algorithms \cite{Shir06nicheradius}. 

\subsection{Multiobjective Approach}
At this point, we would like to note that the readers should not confuse evolutionary \textbf{multimodal} optimization (main theme of this chapter) with evolutionary \textbf{multiobjective} optimization. The former aims at solving  a single function for multiple opima, while the latter aims at solving multiple functions for pareto front solutions. Nonetheless, the techniques involved are related. In particular, Deb and Saha demonstrated that, by decomposing a single multimodal objective function problem into a bi-objective problem, they can solve a multimodal function using a evolutionary multiobjective optimization algorithm \cite{Deb:2010:FMS:1830483.1830568}. Briefly, they keep the original multimodal objective function as the first objective. On the other hand, they use the gradient information to define peaks in the second objective.

\subsection{Ensemble}

As mentioned in the previous section, different niching algorithms have been proposed over the past years. Each algorithm has its own characteristics and design philosophy behind. Although it imposes difficult conditions to compare them thoroughly, it is a double-edged sword. Such a vast amount of algorithms can provide us a ``swiss army knife'' for optimizations on different problems. In particular, Yu and Suganthan proposed an ensemble method to combine those algorithms and form a powerful method called Ensemble of Niching Algorithms (ENA) \cite{Yu:2010:ENA:1808358.1808717}.  An extension work can also be found in \cite{5586341}.

\subsection{Others}
Researchers have been exploring many different ways to deal with the problem. Those methods include: clearing \cite{petrowski:clearing}, repeated iterations \cite{beasley:iteration}, species-specific explosion \cite{kcwong:EASE}, traps \cite{karatsu:globally}, stochastic automation \cite{MOsearch}, honey bee foraging behavior \cite{MObee}, dynamic niching \cite{MOPSO}, spatially-structured clearing \cite{5586085}, cooperative artificial immune network \cite{4630998}, particle swarm optimization \cite{4424892,5352335,Juang2010,Liu2010} , and island model \cite{bessaou:island}. In particular, Stoean et al. have proposed a topological species conservation algorithm in which the proper topological separation into subpopulations has given it an advantage over the existing radius-based algorithms \cite{5491155}. Comparison studies were conducted by Singh et al. \cite{singh:comparison}, Kronfeld et al. \cite{5585916}, and Yu et al. \cite{4631090}. Though different methods were proposed in the past, they are all based on the same fundamental idea: it is to strike an optimal balance between convergence and population diversity in order to locate multiple optima simultaneously in a single run \cite{MOSMC,kcwong:locality}.

\section{Benchmarking}

\subsection{Benchmark Functions}
There are many multimodal functions proposed for benchmarking in the past literature. In particular, the following five benchmark functions are widely adopted in literature: Deb's 1st function \cite{kcwong:EASE}, Himmelblau function \cite{beasley:iteration}, Six-hump Camel Back function \cite{test:SHCB}, Branin function \cite{test:SHCB}, and Rosenbrock function \cite{test:Rosenbrock}. In addition, five more benchmark functions (PP1 to PP5) can be found in \cite{kcwong:EASE,rodica:2008}. For more viogorous comparisons, the IEEE Congress on Evolutionary Computation (CEC) usually releases a test suite for multimodal optimization every year. More than 15 test functions can be found there \cite{Li13benchmarkfunctions}.

\subsection{Performance Metrics}
Several performance metrics have been proposed in the past literature \cite{rodica:2008,li:SDE,li:SCGA,thomsen:crowdingde}. Among them, Peak Ratio (PR) and Average Minimum Distance to the Real Optima (D) \cite{rodica:2008,kcwong:EASE} are commonly adopted as the performance metrics. 

\begin{itemize}
\item A peak is considered found when there exists a individual which is within 0.1 Euclidean distance to the peak in the last population. Thus the Peak Ratio is calculated using the equation (3):
\begin{equation*}
Peak\ Ratio = \frac{Number\:of\:peaks\:found}{Total\:number\:of\:peaks}
\tag{3}
\end{equation*}
\item The average minimum distance to the optima (D) is calculated using the equation (4):
\begin{equation*}
D = \frac{\displaystyle\sum_{i=1}^{n} \min_{indiv \in pop}{d(peak_{i},indiv)}}{n}
\tag{4}
\end{equation*}
where $n$ is the number of peaks, $indiv$ denotes a individual, $peak_{i}$ is the $i$th peak, $pop$ denotes the last population, and $d(peak_{i},indiv)$ denotes the distance between $peak_{i}$ and $indiv$.
\end{itemize}
As different algorithms perform different operations in one generation, it is unfair to set the termination condition as the number of generations. Alternatively, it is also unfair to adopt CPU time because it substantially depends on the implementation techniques for different algorithms. For instance, the sorting techniques to find elitists and the programming languages used. In contrast, fitness function evaluation is always the performance bottleneck \cite{NFEexample} \footnote{For instance, over ten hours are needed to evaluate a calculation in computational fluid dynamics \cite{FitnessEvaluations2}}. Thus the number of fitness function evaluations is suggested to be adopted as the running or termination condition for convergence analysis.

\subsection{Statistical Tests}
Since evolutionary multimodal optimization is stochastic in nature, multiple runs are needed to evaluate each method on each test function. The means and standard deviations of performance metrics are usually reported for fair comparison. To justify the results, statistical tests are usually adopted to assess the statistical significances. For instance, t-tests, Mann-Whitney U-tests (MWU), and Kolmogorov-Smirnov test (KS).

\section{Application}

Holographic gratings have been widely used in optical instruments for aberration corrections. In particular, Varied-Line-Spacing (VLS) holographic grating is distinguished by the high order aberration eliminating capability in diffractive optical systems. It is commonly used in high resolution spectrometers and monochromaters. A recording optical system of VLS holographic grating is outlined in \cite{opticalApp}. 

\subsection{Problem Modelling}
The core component descriptions of the optical systems are listed as follows \cite{opticalApp2005}: 
\begin{tabbing}
\=$M_{1},M_{2}$ \=: Two spherical mirrors \\
\>$C,D$        	\>: Two coherent point sources \\
\>$G$ 							\>: A grating blank \\
\end{tabbing}
In this system, there are two light point sources $C$ and $D$. They emit light rays which are then reflected by mirrors $M_{1}$ and $M_{2}$ respectively. After the reflection, the light rays are projected onto the grating blank $G$. More details are given in \cite{opticalApp,opticalApp1995}.
The objective for the design is to find several sets of design variables (or recording parameters \cite{opticalApp}) to form the expected groove shape of $G$ (or the distribution of groove density \cite{opticalApp2005}). 
The design variables are listed as follows:
\begin{tabbing}
\=$\gamma$ \hspace{2pt} \=: The incident angle of the ray $O_{1}O$ \\
\>$\eta_{C}$ \>: The incident angle of the ray $CO_{1}$ \\
\>$\delta$ \>: The incident angle of the ray $O_{2}O$ \\
\>$\eta_{D}$ \>: The incident angle of the ray $DO_{2}$ \\
\>$p_{C}$ \>: The distance between $C$ and $M_{1}$ ($CO_{1}$) \\
\>$q_{C}$ \>: The distance between $M_{1}$ and $G$ ($O_{1}O$) \\
\>$p_{D}$ \>: The distance between $D$ and $M_{2}$ ($DO_{2}$) \\
\>$q_{D}$ \>: The distance between $M_{2}$ and $G$ ($O_{2}O$) \\
\end{tabbing}
Mathematically, the goal is to minimize the definite integral of the square error between the expected groove density and practical groove density \cite{opticalApp}:
\(
min \hspace{6pt} J = \int^{w_{0}}_{-w_{0}} (n_{p}-n_{e})^{2} dw
\)
where $w_{0}$ is the half-width of the grating, $n_{p}$ is the practical groove density, and $n_{e}$ is the expected groove density. These two groove densities are complicated functions of the design variables. 
Ling et al. have further derived the above formula into a simpler one \cite{opticalApp2005}:
\[
min \hspace{6pt} J = r_{1}^{2} + \frac{w_{0}^{2}(2r_{1}r_{3}+r_{2}^{2})}{3} + \frac{w_{0}^{4}(r_{3}^{2}+2r_{2}r_{4})}{5} + \frac{w_{0}^{6}r_{4}^{2}}{7}
\]
\[
r_{1} = \frac{j_{10}}{\lambda_{0}} - n_{0}\ ,\ \ \ \ \ \ \ \ \ \ r_{2} = \frac{j_{20}}{\lambda_{0}} - n_{0}b_{2}
\]
\[
r_{3} = \frac{3j_{30}}{2\lambda_{0}} - n_{0}b_{3}\ ,\ \ \ \ r_{4} = \frac{j_{40}}{2\lambda_{0}} - n_{0}b_{4}
\]
where $j_{10}$, $j_{20}$, $j_{30}$ and $j_{40}$ are the functions of the design variables, which are $n_{10}$, $n_{20}$, $n_{30}$ and $n_{40}$ respectively in \cite{opticalApp1995}.
Theoretically, the above objective is simple and clear. Unfortunately, there are many other auxiliary optical components in practice, which constraints are too difficult to be expressed and solved in mathematical forms. An optimal solution is not necessarily a feasible and favorable solution. Optical engineers often need to tune the design variables to find as many optimal solutions as possible for multiple trials. Multimodal optimization becomes necessary for this design problem.

\subsection{Performance measurements}
As the objective function is an unknown landscape, the exact optima information is not available. Thus the previous performance metrics cannot be adopted. We propose two new performance metrics in this section. The first one is the best fitness, which is the fitness value of the fittest individual in the last population. The second one is the number of distinct peaks, where a distinct peak is considered found when there exists a individual which fitness value is below a threshold 0.0001 and there isn't any other individual found as a peak before within 0.1 Euclidean distance in the last population. The threshold is chosen to 0.0001 because the fitness values of the solutions found in \cite{opticalApp} are around this order of magnitude. On the other hand, the distance is chosen to 0.1 Euclidean distance because it has already been set for considering peaks found in peak ratio \cite{kcwong:EASE,rodica:2008}. Nonetheless, it is undeniable that such a threshold may not be suitable for this application because the landscape is unknown, although the value of 0.1 is the best choice we can adopt in this study.

\subsection{Parameter setting}
CrowdingDE-STL \cite{wong2012evolutionary}, CrowdingDE-TL \cite{wong2012evolutionary},  CrowdingDE-SL \cite{wong2012evolutionary}, Crowding Genetic Algorithm (CrowdingGA) \cite{jong:crowding}, CrowdingDE \cite{thomsen:crowdingde}, Fitness Sharing Genetic Algorithm (SharingGA) \cite{goldberg:sharing}, SharingDE \cite{thomsen:crowdingde}, Species Conserving Genetic Algorithm (SCGA) \cite{li:SCGA}, SDE \cite{li:SDE}, and UN \cite{dejong:book} are selected for illustrative purposes in this application. All the algorithms were run up to a maximum of 10000 fitness function evaluations. The above performance metrics were obtained by taking the average and standard deviation of 50 runs. The groove density parameters followed the setting in \cite{opticalApp}: $n_{0} = 1.400 \times 10^{3} (line/mm)$, $b_{2} = 8.2453 \times 10^{-4} (1/mm)$, $b_{3} = 3.0015 \times 10^{-7} (1/mm^2)$ and $b_{4} = 0.0000 \times 10^{-10} (1/mm^3)$. The half-width $w_{0}$ was 90mm. The radii of spherical mirrors $M_{1}$ and $M_{2}$ were 1000mm. The recording wavelength ($\lambda_{0}$) was 413.1nm. The population size of all the algorithms was set to 50. The previous settings remained the same, except the algorithm-specific parameters: The species distance of SDE and SCGA was set to 1000. The scaling factor and niche radius of SharingDE and SharingGA were set to 1 and 1000 respectively. The discount factor of the temporal locality was set to 0.5. The survival selection method of the non-crowding algorithms was set to binary tournament \cite{dejong:book}.

\begin{figure}[ht!]
  \centering
  \subfigure[Best Fitness]{\label{fig:RW1stat}\includegraphics[width=0.48\textwidth]{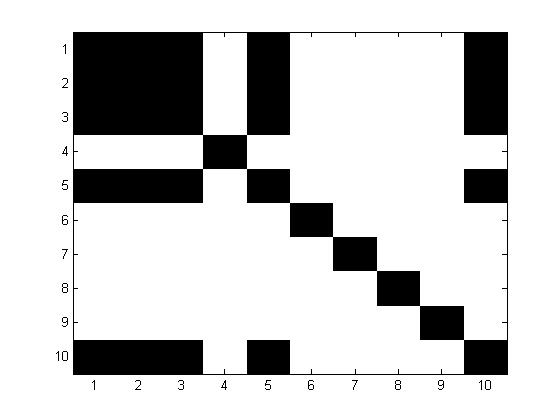}}                
  \subfigure[Best Fitness]{\label{fig:RW2stat}\includegraphics[width=0.48\textwidth]{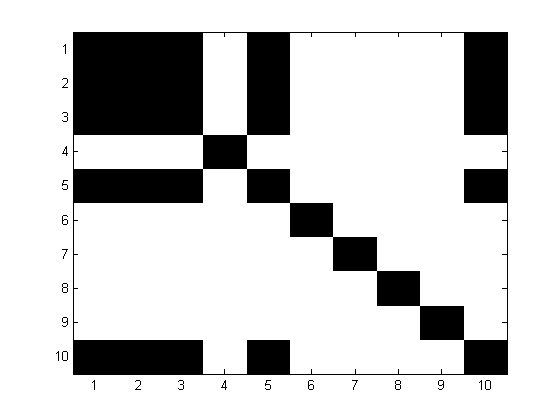}}
  \subfigure[Peaks Found]{\label{fig:RW3stat}\includegraphics[width=0.48\textwidth]{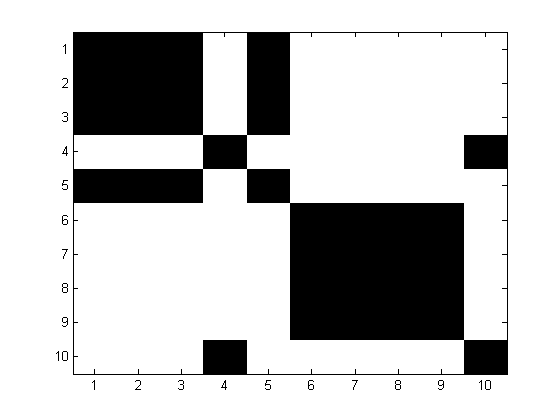}}
  \subfigure[Peaks Found]{\label{fig:RW4stat}\includegraphics[width=0.48\textwidth]{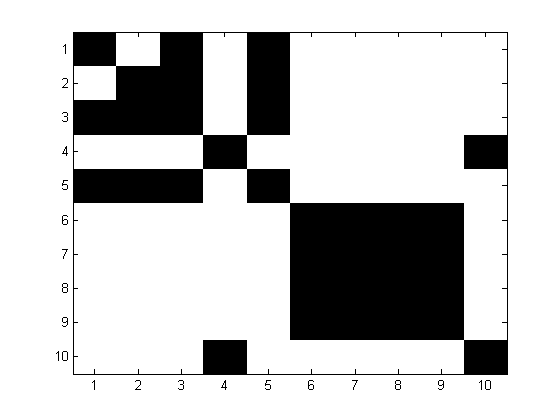}}
  \caption{For Table \ref{tab:RW10000}, we depict the statistical significance test results for the pairwise performance differences between all algorithms tested on the VLS holgraphic grating design problem by Mann-Whitney U-test (MWU) and Two-sample Kolmogorov-Smirnov test (KS) with p=0.05. Each sub-figure correspond to the performance comparison using a metric by a statistical test. The vertical axis is the same as the horizontal axis. Each algorithm is represented by a number on each axis. The numbering of the algorithms follows the order in Table \ref{tab:RW10000}. For instance, 1 refers to CrowdingDE-STL, 2 refers to CrowdingDE-TL......10 refers to UN. The color of each block represents whether the algorithm indicated by the horizontal axis shows a performance different from the algorithm indicated by the vertical axis in a statistically significant way. The black color denotes the p-values higher than 0.05 whereas the white color denotes the p-values lower than 0.05. The even numbered sub-figures are the results obtained by MWU, whereas the odd numbered sub-figures are the results obtained by KS.}
  \label{fig:RWstat}
\end{figure}

\begin{figure}[ht!]
  \centering
  \subfigure[]{\label{fig:RS1}\includegraphics[width=0.7\textwidth]{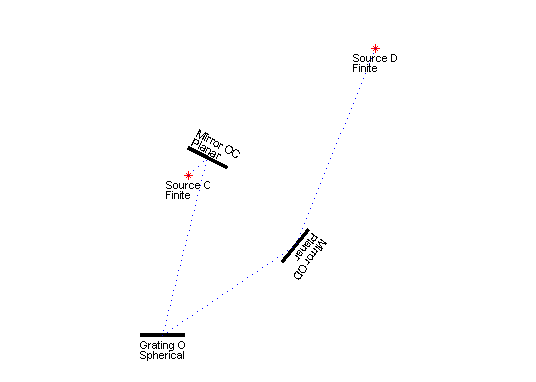}}                
  \subfigure[]{\label{fig:RS2}\includegraphics[width=0.7\textwidth]{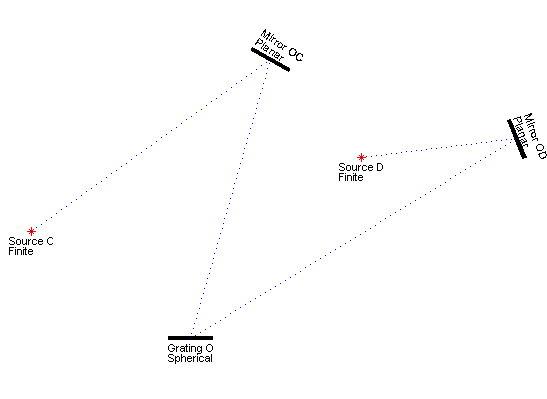}}
  \caption{Configurations obtained by a single run of CrowdingDE-STL on the VLS holographic grating design problem. It can be seen that they are totally different and feasible configurations with which optical engineers can feel free to perform multiple trials after the single run.}
  \label{fig:RS}
\end{figure}

\subsection{Results}
The result is tabulated in Table~\ref{tab:RW10000}. It can be observed that  CrowdingDE-STL can achieve the best fitness whereas CrowdingDE-TL can acheive the best number of peaks found. To compare the algorithms rigorously, statistical tests have also been used. The results are depicted in Figure \ref{fig:RWstat}. One can observe that there are some statistically significant performance differences among them. In particular, CrowdingDE-based methods are shown to have the results statistically different from CrowdingGA, SharingGA, SharingDE, SDE, and SCGA. Some configurations obtained after a run of CrowdingDE-STL on this problem are depicted in Figure \ref{fig:RS}. It can be seen that they are totally different and feasible configurations with which optical engineers can feel free to perform multiple trials after the single run.

\begin{table*}
  \caption{Results for all algorithms tested on the VLS holographic grating design problem (50 runs)}
\centering
\resizebox{\textwidth}{!} {
\begin{tabular}{l*{5}{c}}
\hline\hline
Measurement & CrowdingDE-STL \cite{wong2012evolutionary} & CrowdingDE-TL \cite{wong2012evolutionary} & CrowdingDE-SL \cite{wong2012evolutionary} & CrowdingGA \cite{jong:crowding} & CrowdingDE \cite{thomsen:crowdingde} \\
\hline\hline
Mean of Best Fitness & \textbf{8.29E-08} & 7.17E-07 & 1.18E-07 & 9.02E-06 & 3.66E-06  \\
StDev of Best Fitness & \textbf{2.91E-07} & 5.01E-06 & 4.04E-07 & 3.40E-05 & 2.31E-05 \\
Mean of Peaks Found & 	41.42 &  \textbf{45.54} & 	43.38 & 	8.94 & 	41.98 \\
StDev of Peaks Found & 13.07  & \textbf{9.00}  & 10.69  & 5.04  & 14.26 \\
\hline\hline
Measurement & SharingGA \cite{goldberg:sharing} & SharingDE \cite{thomsen:crowdingde} & SDE \cite{li:SDE} & SCGA \cite{li:SCGA} & UN \cite{dejong:book}\\
\hline\hline
Mean of Best Fitness & 1.87E+04 & 1.74E+02 & 1.13E+00 & 1.24E+02 & 9.19E-04\\
StDev of Best Fitness & 6.82E+04 & 2.65E+02 & 1.56E+00 & 4.59E+02 & 3.15E-03\\
Mean of Peaks Found & 	0.0 & 	0.0 & 	0.06 & 	0.02 & 	7.22\\
StDev of Peaks Found  & 	0.0 & 	0.0 & 0.31  & 0.14  & 3.92 \\
\hline
\end{tabular}
}
\label{tab:RW10000}
\end{table*}

\section{Discussion}
To conclude, we have briefly reviewed the state-of-the-arts methods of evolutionary multimodal optimization from different perspectives in this chapter. Different evolutionary multimodal optimization methodologies are described. To compare them fairly, we described different benchmarking techniques such as performance metrics, test functions, and statistical tests. An application to Varied-Line-Spacing (VLS) holographic grating is presented to demonstrate the real-wold applicability of evolutionary multimodal optimization. Nonetheless, we would like to note several current limitations of evolutionary multimodal optimization as well as the possible solutions at the end of this chapter.

First, most of the past studies just focus on low dimensional test functions for benchmarking. More high dimensional test functions should be incorporated in the future. Second, we would like to point out that evolutionary multimodal optimization is actually far from just finding multiple optima because the algorithms for multimodal optimization usually not only locate multiple optima in a single run, but also preserve their population diversity throughout a run, resulting in their global optimization ability on multimodal functions. Moreover, the techniques for multimodal optimization are usually borrowed as diversity maintenance techniques to other problems. Third, the computational complexities of the methods are usually very high comparing with the other methods since they involve population diversity maintenance which implies that the related survival operators need to take into account the other individuals, resulting in additional time complexity.

\bibliographystyle{plain}
\bibliography{bibMO}

\end{document}